\documentclass{article}

\usepackage[accepted]{icml2024}

\makeatletter
\renewcommand{\ICML@appearing}{Preprint. Work in progress.}
\makeatother

\usepackage[utf8]{inputenc}
\usepackage[T1]{fontenc}
\usepackage{hyperref}
\usepackage{url}
\usepackage{graphicx}
\usepackage{booktabs}
\usepackage{amsmath}
\usepackage{amssymb}
\usepackage{microtype}
\usepackage{caption}
\begin{document}

\twocolumn[
\icmltitle{ComplexityWorld: Benchmarking Vision-Language Models on\\ Verifiable Visual Decision Making}


\icmlsetsymbol{equal}{*}

\begin{icmlauthorlist}
  \icmlauthor{Ningxin Pan}{a}
  \icmlauthor{Hanyu Li}{b}
  \icmlauthor{Yehui Tang}{c}
\end{icmlauthorlist}

\icmlaffiliation{a}{Peking University, Beijing, China. \texttt{pnx20030929@stu.pku.edu.cn}}
\icmlaffiliation{b}{Peking University, Beijing, China. \texttt{lhydave@pku.edu.cn}}
\icmlaffiliation{c}{Samsung Research, Beijing, China. \texttt{yehui.tang@samsung.com}}

\icmlcorrespondingauthor{Yehui Tang}{yehui.tang@samsung.com}

\vskip 0.3in
]

\printAffiliationsAndNotice{}

\begin{abstract}
Vision-language models (VLMs) have made rapid progress in visual perception and increasingly support real-world tasks that depend on images. Many such tasks, however, require more than recognizing what an image contains: a model must use visual evidence to make a complete decision whose parts jointly satisfy global constraints. We introduce \textsc{ComplexityWorld}, a benchmark of 390 tasks across 39 domain-inspired visual worlds and 29 decision categories. Each task is generated from a hidden structured specification, rendered as a visual scene, and scored by an executable verifier that accepts any feasible solution. Under direct inference, all evaluated models except GPT-5.6-Sol remain below 40\% verifier acceptance rate (VAR), while GPT-5.6-Sol reaches 75.6\%. Performance improves substantially when the same decision information is made explicit in structured form, yet varies sharply across equivalent visual presentations. Agent scaffolds provide smaller, model-dependent gains. Together, these results reveal a persistent visual-to-decision bottleneck that additional inference alone does not remove.
\end{abstract}

\section{Introduction}

Visual perception has long been a central goal for vision-language models (VLMs). Modern VLMs can recognize objects, read text, describe scenes, and answer questions about images, supporting a broad range of real-world needs, including document understanding, chart analysis, and visual assistance. Yet as these models move from interpreting visual content to acting on it, perception is no longer enough. They must use visual evidence to make complete decisions.

Here, a decision is not a label or a single action. It is a structured solution whose components must jointly satisfy the constraints shown in the image. A route must respect every visible closure and connection; a schedule must combine displayed dependencies with resource limits; and a spatial arrangement must obey both geometry and obstacles. In each case, the image is the primary source of decision evidence rather than optional context, and one missed relation can invalidate an otherwise reasonable solution.

Existing benchmarks measure many prerequisites for this capability, including visual question answering, mathematical reasoning, abstraction, puzzles, planning, and optimization \citep{masry2022chartqa,lu2024mathvista,chen2026babyvision,cai2025mmiq,ren2025vgrp,zhang2025puzzlebench,mayer2025ivispar,ji2025mpcc,li2026mmopt}. However, they often end with a short answer, use a fixed visual language, cover only a few environments, or provide the task structure in an already organized form. It therefore remains unclear whether a VLM can recover a decision problem from a diverse visual scene and construct a complete solution that satisfies all of its constraints.

We introduce \textsc{ComplexityWorld} to evaluate this missing capability. Its frozen main panel contains 390 tasks spanning 29 decision categories and 39 domain-inspired visual worlds, expressed through maps, diagrams, boards, grids, and spatial layouts. Given an image, instruction, and structured output format, a model must recover the relevant entities and relations and return a complete decision rather than a label or intermediate step.

ComplexityWorld uses a generate-render-verify pipeline to make these decisions both diverse and exactly measurable. A seeded generator first creates a solvable problem with known entities, relations, and constraints. A domain adapter then renders that problem using domain-specific names, icons, and layouts. Models see only the image, instruction, and answer format; the underlying problem record, construction solution, and verifier remain hidden. After inference, a task-specific program checks every constraint and accepts any feasible solution. The same problem can also be rendered in different visual forms, enabling controlled tests of presentation sensitivity.

The results expose a clear visual-to-decision bottleneck. Under direct inference, GPT-5.6-Sol reaches 75.6\% VAR, while Qwen3.7-Plus, Gemini-3.5-Flash, and MiMo-v2.5 reach 39.5\%, 32.3\%, and 24.4\%, respectively. Replacing images with answer-free explicit-structure records improves Qwen3.7-Plus by 40.6 points and GPT-5.6-Sol by 21.4 points, while equivalent problems rendered in different visual forms produce gaps of up to 38.3 points. Thus, reliably organizing visual evidence is itself a major challenge, beyond constructing a solution once the problem is explicit.

Finally, as a secondary comparison, we test two model-matched agent scaffolds. Codex improves GPT-5.6-Sol by 7.4 points and MiMo Code improves MiMo-v2.5 by 3.8 points, indicating smaller and model-dependent gains than the representation diagnostics.

\begin{figure*}[t]
\centering
\includegraphics[width=0.98\textwidth]{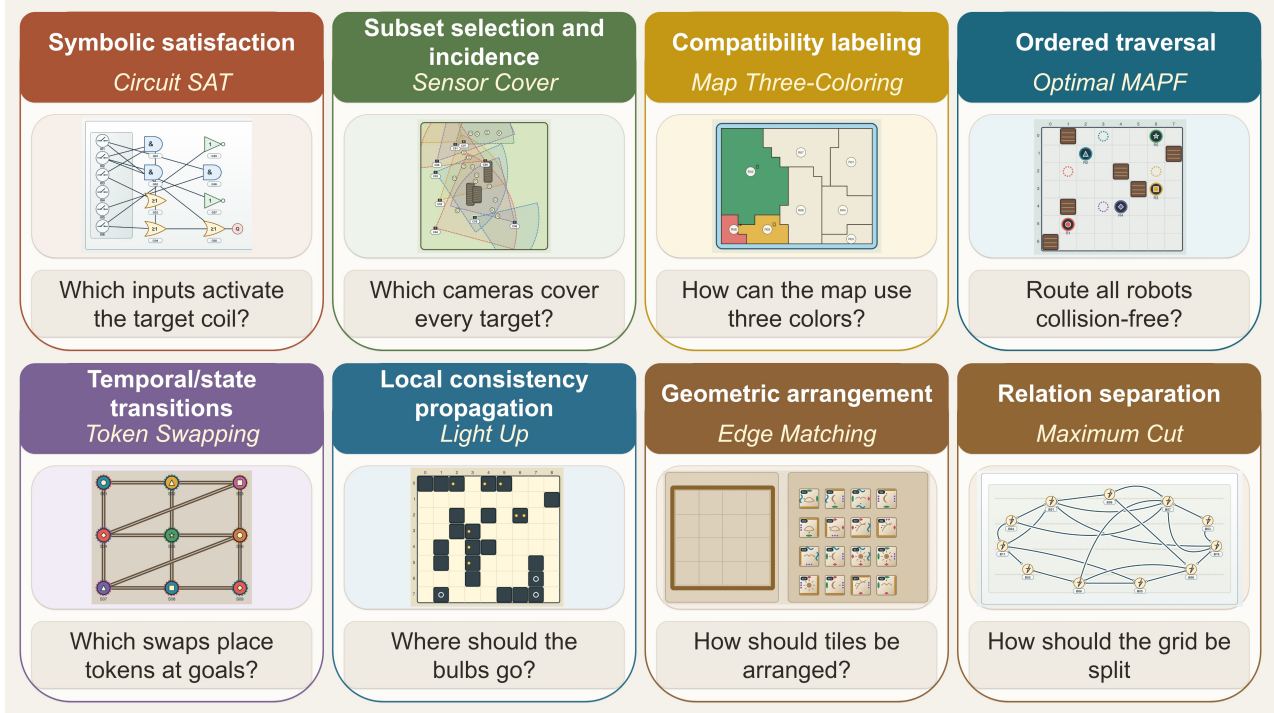}
\caption{Representative ComplexityWorld tasks organized by eight constraint-signature families. These family labels summarize the dominant form of their constraints and the structure of the output checked by the verifier, whereas the 29 decision categories name task semantics or application operations. A model reads each rendered scene and returns a complete structured decision checked by an executable verifier.}
\label{fig:worlds}
\end{figure*}

Our contributions are threefold:
\begin{itemize}
    \item We introduce ComplexityWorld, a 390-task benchmark for complete visual decisions across 39 visual worlds and 29 decision categories.
    \item We develop a generate-render-verify pipeline that supports diverse visual forms, multiple feasible outputs, and exact constraint-based evaluation.
    \item We identify a large visual-to-decision bottleneck through explicit-structure and matched-presentation studies, then use agent scaffolds as a secondary test of how much additional inference can recover.
\end{itemize}

\section{Related Work}

\paragraph{From visual understanding to decision construction.}
Visual reasoning benchmarks span controlled scenes, expert questions, charts, diagrams, mathematical reasoning, and abstraction \citep{johnson2017clevr,yue2024mmmu,masry2022chartqa,singh2024flowvqa,lu2024mathvista,cai2025mmiq,chen2026babyvision,yue2024mmmupro}. They establish important perceptual and reasoning capabilities, but usually end with a label, option, number, or another localized response. ComplexityWorld instead requires a complete solution whose fields jointly satisfy constraints recovered from the image. The distinction is therefore not merely answer length, but inferring a local conclusion versus constructing a globally valid decision.

\paragraph{Visual puzzles, planning, and optimization.}
Generated puzzles, visual planning tasks, and multimodal optimization benchmarks add controlled generation, multi-step constraints, or executable evaluation \citep{fan2024nphard,ren2025vgrp,zhang2025puzzlebench,waugh2026pencil,mayer2025ivispar,ji2025mpcc,wang2026worldtravel,li2026mmopt,li2026oragent}. Existing tasks typically focus on a small set of visual languages, interactive trajectories, intermediate puzzle states, or the construction of a formal model and solver. ComplexityWorld instead spans heterogeneous visual worlds while keeping the evaluation target fixed: recover the decision problem from the image and directly return a complete solution accepted by a hidden verifier.

\paragraph{Executable evaluation and representation controls.}
Programmatic generation and verification enable fresh instances, global constraint checking, and evaluation beyond one serialized reference answer \citep{fan2024nphard,zhang2025puzzlebench,waugh2026pencil,li2026mmopt,li2026oragent}. Prior work also finds sensitivity to input modality and visual presentation \citep{fan2024nphard,colan2025pathplanning,qiu2022multimodalrobustness,yue2024mmmupro}. ComplexityWorld builds on these foundations with two matched controls: an answer-free explicit-structure record preserves the decision and verifier while removing the image, and independently designed renderings preserve the underlying rules while changing how their evidence is organized. Agent methods provide a complementary way to add tools, interaction, and computation \citep{yao2023react,yang2023mmreact,suris2023vipergpt,hu2024visualsketchpad}; we evaluate complete model-matched scaffolds only as a secondary inference condition.

\section{ComplexityWorld Benchmark}

\subsection{Task Interface and Evaluation}

Each task has a public model interface and a private evaluation record. The model sees a rendered image \(x\), a natural-language instruction \(q\), and an output specification \(\Sigma\) defining the required JSON fields, and it returns one complete structured decision \(\hat y\). The evaluator retains the hidden structured task specification \(z\) and task-specific verifier \(V\); the generation-time construction solution is used to establish solvability but is not the scoring target. An answer is correct exactly when it parses under \(\Sigma\), refers to valid task entities, and \(V(z,\hat y)\) confirms that every required constraint is satisfied. Thus, any feasible decision is accepted, even if it differs from the construction solution.

The primary metric is verifier acceptance rate (VAR):
\begin{equation}
\mathrm{VAR}=\frac{1}{N}\sum_{i=1}^{N}
\mathbf{1}\!\left[V_i(z_i,\hat y_i)=1\right].
\label{eq:valid}
\end{equation}

All evaluated instances are solvable and checked during generation. ComplexityWorld therefore evaluates construction of globally valid decisions from visual evidence, rather than infeasibility detection or agreement with one serialized reference answer.

\subsection{Benchmark Taxonomy}

ComplexityWorld is organized at several levels. A \emph{task template} defines an abstract problem generator, output schema, solver, and verifier, while a \emph{visual-world adapter} renders such a problem using domain-specific semantics, icons, labels, and layouts. A selected template--world pair defines one benchmark task type, from which multiple seeded instances are generated. We group templates into 29 semantic \emph{decision categories}, which name the decision being made, and separately summarize them using eight \emph{constraint-signature families}, which capture the dominant constraints and output structure checked by the verifier. Categories may contain multiple task types; families do not alter task sampling or scoring and are not independently validated dimensions of model capability.

The benchmark targets decision categories that recur across many real-world domains. Limited resources may need to cover demands, assignments may need to respect capacities, actions may need to follow dependencies, and objects may need to be arranged without conflict. Task templates involving coverage, for example, can represent facility placement, wildlife monitoring, or emergency communication. Well-studied combinatorial problems formalize these categories independently of one application; rendering them as different visual worlds provides broad coverage without claiming to reproduce every aspect of deployment.

These decision categories also have a useful property for evaluation. Constructing a valid solution requires coordinating choices across the entire input: one route segment, placement, or assignment can restrict decisions elsewhere. Once a model proposes a complete answer, however, a program can check every constraint directly. Many classical problems with this pattern are NP-complete \citep{karp1972reducibility,garey1979computers}, but ComplexityWorld relies only on the intuitive distinction between difficult global construction and efficient verification.

Figure~\ref{fig:worlds} visualizes this hierarchy through representative tasks arranged by the eight constraint-signature families:

\begin{itemize}
    \item \emph{Symbolic satisfaction} recovers Boolean or discrete assignments that satisfy a formula. In the Circuit SAT example, the model assigns truth values to PLC inputs so that the target output coil is energized.
    \item \emph{Subset selection and incidence} selects items subject to coverage, domination, compatibility, or identification constraints. Sensor Cover, for example, asks for a limited set of cameras whose unobstructed fields of view cover every target.
    \item \emph{Relation separation and acyclicity} removes or partitions relations to enforce cuts or eliminate cycles. Examples include partitioning a power network so enough links cross the cut and removing selected vertices from a directed graph to break every cycle.
    \item \emph{Compatibility labeling} assigns mutually compatible labels or correspondences. In Map Three-Coloring, every region receives a label while regions sharing a boundary must receive different labels.
    \item \emph{Ordered traversal} returns a sequence or a set of connected trajectories satisfying traversal constraints. Optimal MAPF requires synchronized robot paths that reach all goals within a time bound without collisions or head-on swaps.
    \item \emph{Temporal/state transitions} produces action sequences under dependencies, resources, or state changes. Token Swapping, for example, requires an ordered sequence of adjacent swaps that moves every token to its destination within a budget.
    \item \emph{Local consistency propagation} combines local clues with board-wide or line-of-sight constraints. In Light Up, bulb placements must satisfy numbered walls, illuminate every open cell, and prevent any two bulbs from seeing each other.
    \item \emph{Geometric arrangement} constructs tilings, packings, or spatial partitions subject to fit and non-overlap constraints. Edge Matching places and rotates square tiles so touching patterns agree, while Tantrix Rotation aligns hexagonal channels and forms a single connected cycle.
\end{itemize}

The panels are examples of the family-level organization, not a one-to-one display of all 29 decision categories. Across categories, families, and visual worlds, the common interface remains the same: recover the relevant constraints from the image and return all mutually dependent choices in the required structured form.

\subsection{Construction Pipeline}

ComplexityWorld separates the machine-readable decision problem from the image shown to the model. This lets us verify solvability and vary the presentation without changing what counts as correct. The construction has four stages: task templates, visual world generation, rendering and quality control, and selection of the frozen evaluation set.

\subsubsection*{Task Templates}

A task template defines a reusable canonical problem specification: its entities and constraints, seeded sampling procedure, required output schema, construction solver, and verifier. To create a task, the generator samples a problem and the solver constructs one valid solution. The verifier must accept that solution before the task is retained. This generation-time answer establishes solvability; it is neither shown to the model nor treated as uniquely correct.

Each task template is assigned to exactly one decision category according to the decision it requires, while a category may contain multiple templates with distinct generators, output schemas, and verifiers.

The template library provides breadth across decision categories. It covers selection, coverage, graph editing, routing, scheduling, packing, coloring, tiling, and state transformation. The evidence needed from the image changes accordingly: routing depends on connectivity and order, packing on geometry and containment, and scheduling on labels and temporal dependencies.

\subsubsection*{Visual-World Adapters}

The task template defines an abstract problem but not how the model will see it. A visual-world adapter supplies that presentation. It maps abstract entities and relations to domain-specific objects, then chooses their displayed names, icons, layout, and instruction.

The adapter produces a model-facing scene description containing only information that may appear in the task. Internal task-family labels, the generation-time solution, solver traces, and verifier code are excluded. For the equivalent-problem visual presentation study, several adapters start from the same underlying decision and verification rules but change the domain semantics, layout, icons, wording, and entity labels shown in the image.

\subsubsection*{Rendering and QC}

 We render each scene as a 1600-by-1000 image and check that the required visual content, labels, legends, and constraints are present. Automated tests reject blank or malformed images, missing entities, disagreements between the scene description and the image, and visible text that reveals an internal task label or answer. Template-specific checks also compare important geometry and relations with the underlying task record. We additionally conduct a blinded check of sampled low-VAR instances to verify that every public image presents its decision-relevant entities, constraints, identifiers, and relations clearly and without ambiguity. The Supplementary Material reports the procedure and reproduces four examples.

\subsubsection*{Frozen Set}

The full generator library contains 113 task templates. The frozen main evaluation set selects 39 template--world pairs, spanning 29 decision categories. Before running any evaluated model, we fixed a deterministic schedule of ten instances per selected template--world pair, yielding \(39\times10=390\) tasks.

The 39 worlds cover 29 decision categories. We release the frozen benchmark---including its manifests, images, output specifications, and verifiers---together with the construction pipeline for all 113 templates. Selection required a complete generator--renderer--verifier path, a readable and self-contained image, material dependence on visual evidence, and successful rendering and leakage checks, while favoring structural and visual coverage.

To stress-test verifier behavior, we use one instance from each of the 113 templates. All 113 generation-time solutions are accepted. We also create 596 altered outputs spanning format, identifier, and decision changes. Every one of the 301 format or identifier changes is rejected; 288 of 295 decision changes are rejected, while seven remain feasible alternatives accepted by the verifier and are provided in the Supplementary Material for inspection. This sensitivity test covers both parsing and constraint checking. Because generation and verification share task-specific code, it provides behavioral evidence rather than an independent correctness validation.

\begin{figure*}[t]
\centering
\includegraphics[width=0.98\textwidth,trim=0 8bp 0 8bp,clip]{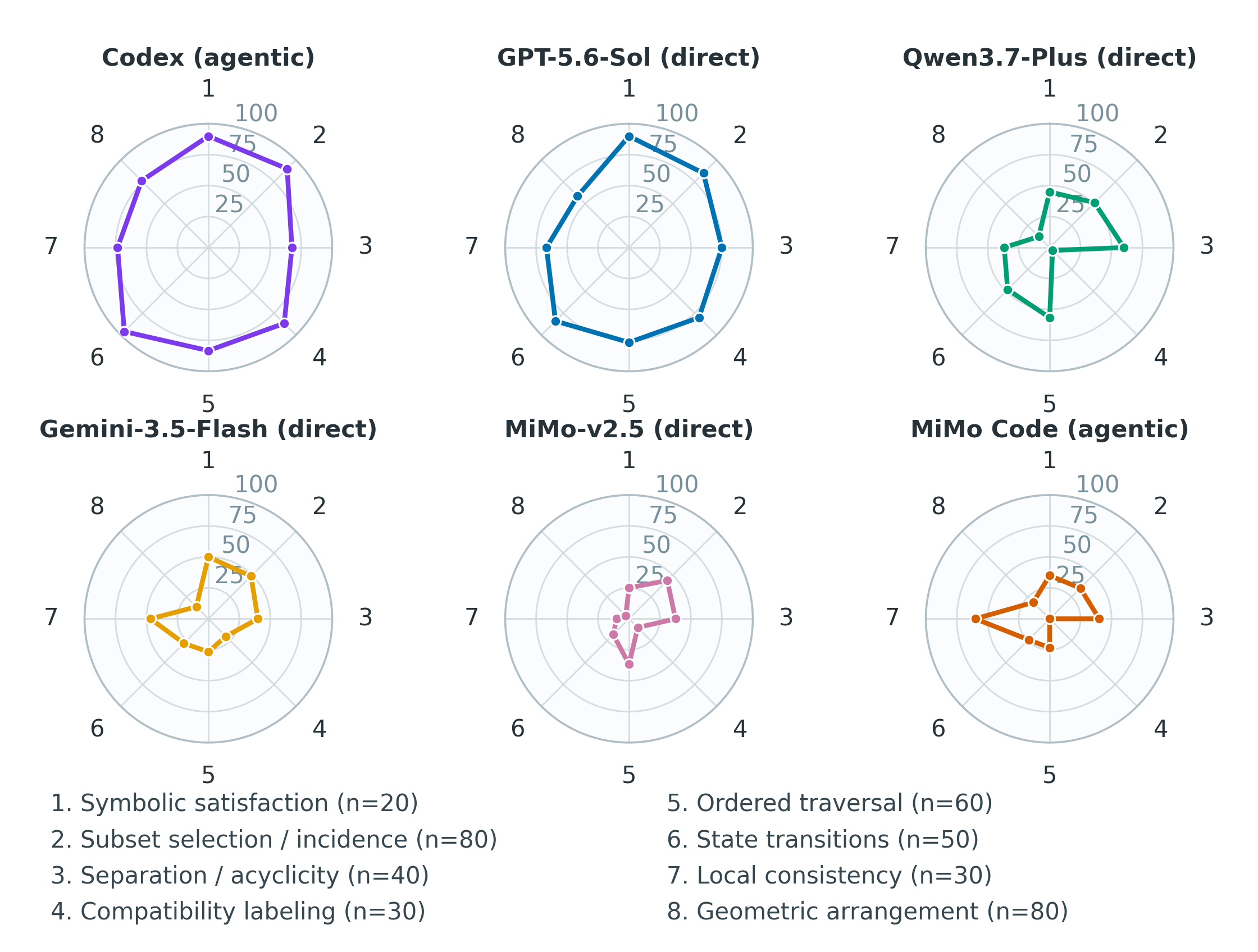}
\caption{Performance profiles across eight constraint-signature families on the frozen evaluation set. Each panel reports one configuration on the full 390-task panel. Lines connect the eight descriptive family scores for readability; enclosed area is not used as a metric.}
\label{fig:capability_radar}
\end{figure*}

\begin{figure*}[!t]
\centering
\includegraphics[width=0.90\textwidth]{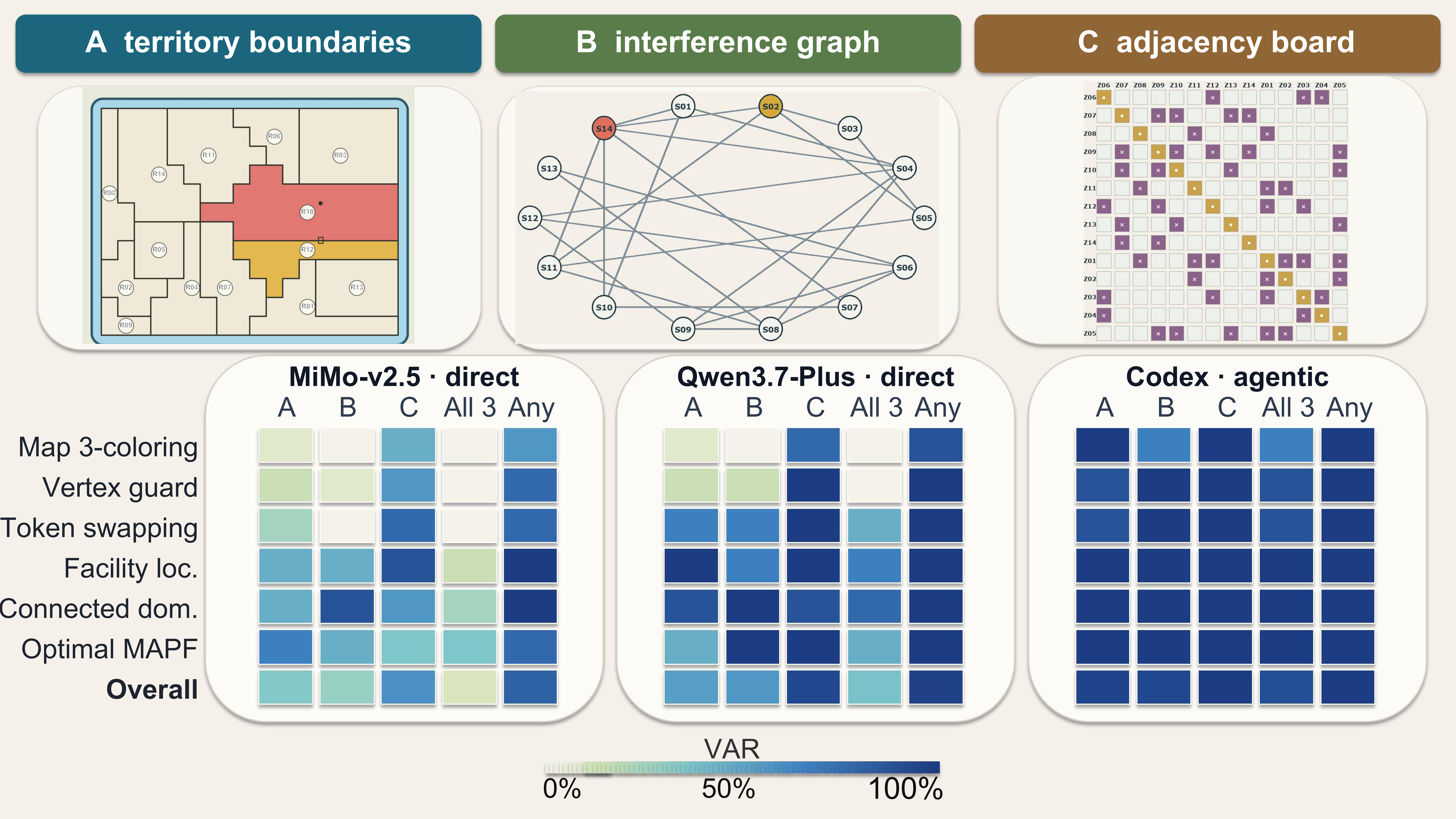}
\caption{Equivalent hidden specifications shown in three visual forms, labeled A, B, and C. \emph{Top:} one map-coloring problem rendered as territory boundaries (A), an interference graph (B), and an adjacency board (C), with the underlying specification, decision type, and verifier held fixed. \emph{Bottom:} VAR across all six task templates in the same A--B--C order. The presentations differ in how the specification is organized; no presentation is assumed to be universally easier.}
\label{fig:paired_models}
\end{figure*}

\section{Experimental Setup}

The primary evaluation compares four VLMs under direct inference: GPT-5.6-Sol, Qwen3.7-Plus, Gemini-3.5-Flash, and MiMo-v2.5. Each model is evaluated on the complete frozen main panel. At inference time, it receives only the image, instruction, and required output fields; the task record and verifier remain hidden until scoring. \emph{Direct} means that the model produces one response without an agent loop or external tools.

The primary metric is VAR (Eq.~\ref{eq:valid}) over one final decision per task. We also report VAR by visual world, decision category, and the eight constraint-signature families. Because every world contributes ten tasks, overall VAR equals the average of per-world VARs. To reduce evaluation cost, the visual-versus-explicit-structure diagnostic uses the same 39 template--world pairs but a separately frozen set of six instances per pair, yielding \(39\times6=234\) diagnostic tasks. Its results are reported separately from the main benchmark.

As a secondary analysis, we evaluate two corresponding agent scaffolds. The scaffolds permit multi-turn interaction, intermediate files, and their native visual and computational tools. For the GPT comparison, Codex wraps the same GPT-5.6-Sol model evaluated in the direct condition; the added components are the scaffold, tools, and additional inference budget. MiMo Code similarly uses MiMo-v2.5 as its underlying model. Agents receive the same public input but no hidden record, solution, or verifier.

The two matched comparisons therefore measure how complete agent setups affect end-to-end performance, with multi-turn interaction, tool access, context limits, and additional computation varying together.

The direct GPT-5.6-Sol run records \texttt{xhigh} reasoning effort; Qwen3.7-Plus and MiMo-v2.5 use \texttt{thinking=auto}; Gemini-3.5-Flash uses the provider-managed \texttt{auto} setting without a separate reasoning-effort override. Available prompts and configuration metadata are provided in the Supplementary Material.

\section{Results and Analysis}

\subsection{Direct VLM Performance}

Table~\ref{tab:direct_results} reports the primary benchmark results. GPT-5.6-Sol reaches 75.6\%, while every other direct VLM remains below 40\%. The 51.2-point gap between GPT-5.6-Sol and MiMo-v2.5 shows that the benchmark separates current models while retaining substantial headroom.

\begin{table}[t]
\centering
{\small
\begin{tabular}{@{}lrr@{}}
\toprule
Direct VLM & Accepted & VAR \\
\midrule
GPT-5.6-Sol & 295/390 & 75.6\% \\
Qwen3.7-Plus & 154/390 & 39.5\% \\
Gemini-3.5-Flash & 126/390 & 32.3\% \\
MiMo-v2.5 & 95/390 & 24.4\% \\
\bottomrule
\end{tabular}
}
\caption{Verifier acceptance rate (VAR) for direct VLM inference on the frozen main panel.}
\label{tab:direct_results}
\end{table}

Aggregate VAR also hides different performance profiles. Qwen3.7-Plus, for example, reaches 51.3--60.0\% on subset-incidence, ordered-traversal, and separation/acyclicity families, but only 3.3\% on compatibility labeling and 12.5\% on geometric arrangement. Figure~\ref{fig:capability_radar} summarizes these differences across all eight constraint-signature families. The variation suggests that failures do not arise from one uniform limitation, motivating the representation diagnostics below.

\subsection{Visual vs.\ Explicit-Structure Diagnostic}

A failure may arise while recovering the decision problem from the image, while constructing a feasible solution, or at both stages. To distinguish these sources, we use the separately frozen 234-instance diagnostic set, with six instances per visual world. The set was fixed before either diagnostic model was run and does not determine the main benchmark score.

\begin{figure}[t]
\centering
\includegraphics[width=0.95\columnwidth]{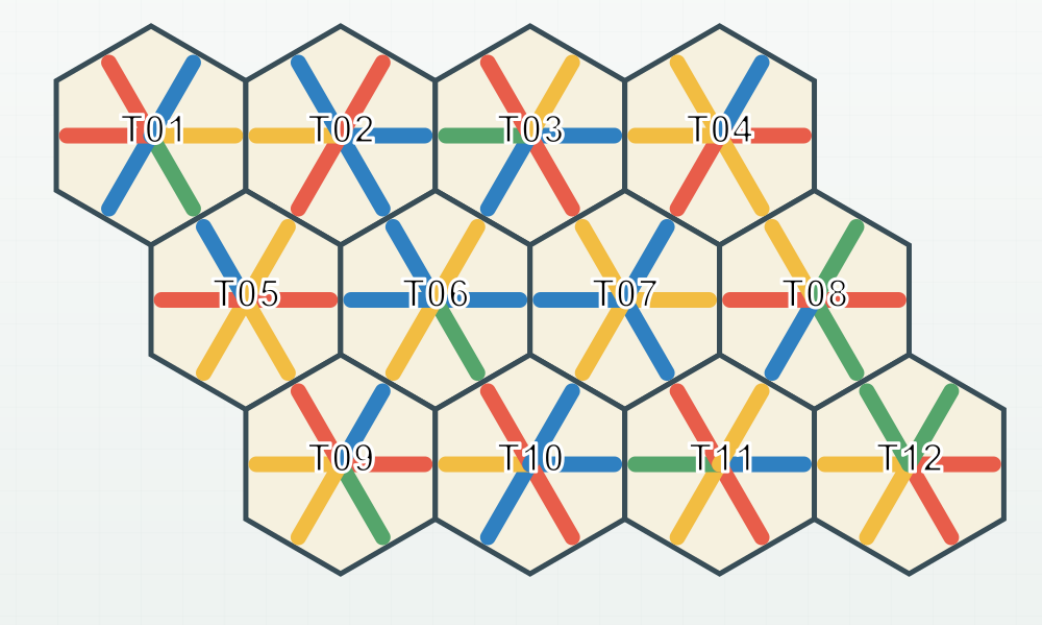}
\caption{Example visual input. One matched answer-free explicit-structure entry is \texttt{\{"id":"T01","row":0,"column":0,"edges":[2,3,1,0,0,1]\}}. It specifies the tile's grid position and six edge-colour indices; the full record contains no answer-bearing information.}
\label{fig:visual-explicit-example}
\end{figure}

For Qwen3.7-Plus and GPT-5.6-Sol, we hold the underlying instances, public instruction, required outputs, completion budget, parser, and verifier fixed. Only the instance representation changes. \emph{Visual} is the standard image condition, whereas the \emph{answer-free explicit-structure record} condition replaces the image with a canonical JSON record of its entities, relations, and constraints. This record uses the same entity names required in the answer but contains no task-family label, reference solution, solver trace, or verifier information. Figure~\ref{fig:visual-explicit-example} illustrates the visual representation and one corresponding record entry.

To confirm that the generic task format does not itself reveal a solution, we also give Qwen only the instruction and output fields, without any instance-specific evidence. It produces no accepted solutions (0/234).

\begin{table}[t]
\centering
{\small
\setlength{\tabcolsep}{3pt}
\begin{tabular}{@{}lrrr@{}}
\toprule
Model & Visual & \shortstack{Explicit-\\structure} & Change \\
\midrule
Qwen3.7-Plus & 88/234 (37.6\%) & 183/234 (78.2\%) & +40.6 \\
GPT-5.6-Sol & 169/234 (72.2\%) & 219/234 (93.6\%) & +21.4 \\
\bottomrule
\end{tabular}
}
\caption{Visual images and answer-free explicit-structure records for the same fixed 234-instance diagnostic subset. Change is in percentage points.}
\label{tab:input_ablation}
\end{table}

Both models improve substantially with answer-free explicit-structure records: Qwen gains 40.6 points and GPT gains 21.4 points. The common direction across models shows that organizing visual evidence into an explicit problem description is a major part of the end-to-end difficulty.

GPT fails 15 tasks in the answer-free explicit-structure record condition, concentrated in feedback-arc-set, LITS, and offline-Tetris instances. Thus, constructing the final output can remain difficult even after the relevant information is explicit.

\subsection{Equivalent-Problem Visual Presentation Study}

VAR on one rendering does not reveal whether a model can recover and solve an equivalent problem in another visual form. We therefore hold each hidden specification and verifier fixed while varying its visual presentation. With six templates and ten problems per template, the study contains 60 matched groups and 180 images per evaluated configuration.

Within each matched group, the underlying entities, constraints, decision type, and verification logic are identical. The three images change the domain framing, layout, icons, wording, and visible entity labels. Presentation C makes pairwise relations explicit: it uses an adjacency, cost, state, or visibility table for five templates and a traversability graph for multi-agent path finding. Presentations A and B preserve more spatial or domain-specific organization, but neither is intended to be universally harder.

\begin{table}[t]
\centering
{\small
\setlength{\tabcolsep}{5pt}
\begin{tabular}{@{}lrrrr@{}}
\toprule
Configuration & A & B & C & All 3 \\
\midrule
Qwen direct & 56.7 & 60.0 & 95.0 & 41.7 \\
MiMo direct & 38.3 & 33.3 & 63.3 & 15.0 \\
Codex agent & 96.7 & 95.0 & 100.0 & 91.7 \\
\bottomrule
\end{tabular}
}
\caption{Verifier acceptance rate (VAR, \%) for equivalent hidden specifications rendered in three visual forms. All 3 is the percentage of underlying problems solved in every presentation. C significantly exceeds A and B for both direct models (Holm-adjusted exact McNemar \(p\leq 0.0271\)).}
\label{tab:paired_summary}
\end{table}

We report VAR for each presentation and \emph{all-three success}, the fraction of problems solved in every form, in Figure~\ref{fig:paired_models} and Table~\ref{tab:paired_summary}. Statistical comparisons preserve matched groups: confidence intervals resample groups within each template, and pairwise differences use exact McNemar tests with Holm correction. Presentation C is strongest overall for both direct models, but success across all three forms remains much lower. Because several visual factors change together, the study measures presentation sensitivity rather than the causal effect of one design choice. Codex is included only as a stronger inference reference.

\subsection{Agent-Scaffolded Inference}
\label{sec:agent}

The preceding studies show that end-to-end VAR changes substantially when the same decision information is reorganized or presented in another form. An agent scaffold may help by allowing iterative image inspection, written intermediate representations, and computational checks. We therefore compare each agent setting with direct inference from the same underlying model on the frozen main panel (Table~\ref{tab:agent_results}).

\begin{table}[t]
\centering
{\small
\begin{tabular}{@{}lrrr@{}}
\toprule
Underlying model & Direct & Agent & Change \\
\midrule
GPT-5.6-Sol & 75.6 & 83.1 & +7.4 \\
MiMo-v2.5 & 24.4 & 28.2 & +3.8 \\
\bottomrule
\end{tabular}
}
\caption{Verifier acceptance rate (VAR, \%) under direct and agent-scaffolded inference on the frozen main panel.}
\label{tab:agent_results}
\end{table}

Codex improves GPT-5.6-Sol by 7.4 points, and its world-stratified 95\% confidence interval excludes zero. MiMo Code is 3.8 points above direct MiMo-v2.5, but its interval includes zero, so the evidence supports only a descriptive gain for this pair. Agent scaffolds can therefore help, but the effect is neither uniform nor separable from the additional tools, interaction, context, and computation supplied by the complete setup.

Complete task-template and decision-category outcomes are provided in the Supplementary Material.

\section{Discussion and Limitations}

ComplexityWorld uses domain-inspired synthetic scenes rather than deployed screenshots, and the main panel contains ten tasks per world. Performance may differ on noisier real-world imagery; more instances and independent renderers would strengthen category-level conclusions. All evaluated problems are solvable, so the benchmark does not test infeasibility detection.

The representation studies identify sensitivity without isolating individual causes. The equivalent-problem study covers six templates and changes layout, iconography, wording, and domain semantics together, while the visual-versus-explicit-structure diagnostic changes both modality and information organization.

The agent comparisons likewise change tools, interaction, context, and computation together. They characterize complete inference configurations rather than the causal effect of any one scaffold component; cross-provider usage is not directly comparable.

\subsection{Implications for Training and Generalization}

Although this paper reports no training experiment, the released generators and verifiers could support training from verified outcomes. Such work should keep all presentations of one underlying problem in the same split, exclude evaluation seeds, and ideally hold out entire visual-world adapters when testing generalization.

\section{Conclusion}

ComplexityWorld tests whether VLMs can recover a decision problem from diverse visual scenes and return a complete solution accepted by an executable verifier. The central finding is not a single leaderboard ordering: success depends strongly on how decision-relevant information is organized and presented. Making the problem explicit through an answer-free explicit-structure record, or visually organizing it to expose key relations, can make the same underlying decision substantially easier. Strong solution construction in one representation therefore does not imply reliable visual decision making across representations.

Agent scaffolds can improve end-to-end performance, but they do not remove this dependence on the input representation. Reliable visual decision making therefore requires both robust recovery of decision-relevant structure and globally consistent construction of the final decision. By releasing the generators, frozen tasks, and executable verifiers, ComplexityWorld provides a reproducible test of these two requirements across heterogeneous visual worlds.

\bibliographystyle{icml2024}
\bibliography{references}

\end{document}